\documentclass{article} 
\usepackage{tencent_ailab_tech_report}
\usepackage[colorlinks = true,
            linkcolor = blue,
            urlcolor  = blue,
            citecolor = blue,
            anchorcolor = blue]{hyperref}   
\usepackage{microtype}
\usepackage{hyperref}
\usepackage{url}
\usepackage{booktabs}
\usepackage{enumitem}
\usepackage{multicol}
\usepackage{multirow}
\usepackage{CJKutf8}
\usepackage{amsmath}
\usepackage{siunitx}
\usepackage{floatflt}
\usepackage{wrapfig}
\usepackage{graphicx}
\usepackage{booktabs}
\usepackage{wrapfig}
\usepackage{authblk}

\usepackage{lipsum}

\usepackage{algorithm}
\usepackage{algorithmicx}
\usepackage{algpseudocode}
\usepackage{microtype}
\usepackage{subfigure}
\usepackage{multirow}
\usepackage{booktabs} 
\usepackage{pifont}  
\usepackage{graphicx}  
\usepackage{subcaption} 

\usepackage{hyperref}

\usepackage{amssymb} 
\DeclareMathOperator{\E}{\mathbb{E}}
\usepackage{amsmath, amssymb}

\algnewcommand{\LeftComment}[1]{\Statex \(\triangleright\) #1}

\usepackage{array}
\usepackage{amsmath}
\usepackage{amssymb}
\usepackage{mathtools}
\usepackage{amsthm}
\usepackage{arydshln}
\usepackage[capitalize,noabbrev]{cleveref}
\usepackage{adjustbox} 
\usepackage{enumitem}
\usepackage{xr}
\usepackage{listings}
\usepackage{bbm}

\theoremstyle{plain}

\theoremstyle{definition}

\theoremstyle{remark}

\usepackage[textsize=tiny]{todonotes}

\definecolor{nred}{RGB}{196, 38, 11}
\definecolor{ngreen}{RGB}{18, 141, 21}
\definecolor{nblue}{RGB}{41, 52, 190}
\definecolor{hzw}{RGB}{223, 97, 76}
\definecolor{lt}{RGB}{54, 89, 170}

\newcommand{\ignore}[1]{}

\colmfinalcopy

\title{{\em Crossing the Reward Bridge:}\\Expanding RL with Verifiable Rewards Across Diverse Domains}

\author[ ]{Yi Su\thanks{The work was done during Yi's internship at Tencent AI Lab.}~~$^{,1,2}$}
\author[ ]{Dian Yu$^{1}$}
\author[ ]{Linfeng Song$^{1}$}
\author[ ]{Juntao Li$^{2}$}
\author[ ]{\mbox{Haitao Mi}$^{1}$}
\author[ ]{\\\mbox{Zhaopeng Tu}$^{1}$}
\author[ ]{\mbox{Min Zhang}$^{2}$}
\author[ ]{Dong Yu$^{1}$}

\affil[1]{Tencent AI Lab}
\affil[2]{Soochow University}

\begin{document}

\maketitle

\begin{figure*}[h!]
   \begin{center}
   \vspace{-20pt}
   \includegraphics[width=0.75\textwidth]{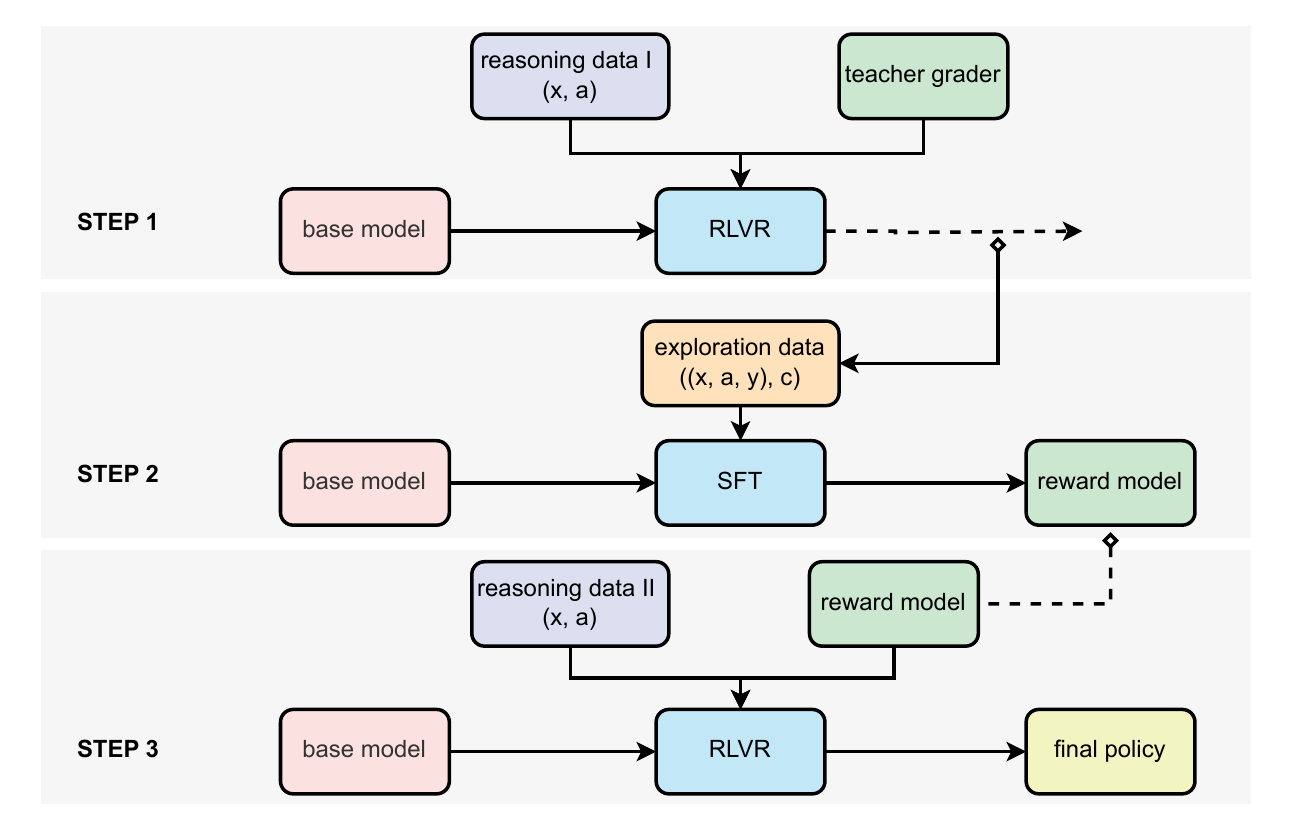}
   \end{center}
 \caption{Overview paradigm of RLVR with our cross-domain verifier.}
 \label{fig:overview}
\end{figure*}

\begin{abstract}

Reinforcement learning with verifiable rewards (RLVR) has demonstrated significant success in enhancing mathematical reasoning and coding performance of large language models (LLMs), especially when structured reference answers are accessible for verification. However, its extension to broader, less structured domains remains unexplored. In this work, we investigate the effectiveness and scalability of RLVR across diverse real-world domains including medicine, chemistry, psychology, economics, and education, where structured reference answers are typically unavailable. We reveal that binary verification judgments on broad-domain tasks exhibit high consistency across various LLMs provided expert-written reference answers exist. Motivated by this finding, we utilize a generative scoring technique that yields soft, model-based reward signals to overcome limitations posed by binary verifications, especially in free-form, unstructured answer scenarios. We further demonstrate the feasibility of training cross-domain generative reward models using relatively small (7B) LLMs without the need for extensive domain-specific annotation. Through comprehensive experiments, our RLVR framework establishes clear performance gains, significantly outperforming state-of-the-art open-source aligned models such as Qwen2.5-72B and DeepSeek-R1-Distill-Qwen-32B across domains in free-form settings. Our approach notably enhances the robustness, flexibility, and scalability of RLVR, representing a substantial step towards practical reinforcement learning applications in complex, noisy-label scenarios.

\end{abstract}

\section{Instruction}

Reinforcement learning with verifiable rewards (RLVR) has recently emerged as an effective paradigm for improving the reasoning capabilities of large language models (LLMs) \citep{luong2024reft, lambert2024t}, even in scenarios without supervised fine-tuning \citep{guo2025deepseek}. RLVR typically leverages reference-based signals, assuming the availability of \textbf{objective ground-truth answers} to determine whether model responses align with reference outcomes. 
In prior studies, RLVR has mainly demonstrated success on tasks with precisely structured solutions, such as mathematical reasoning or code generation, where binary verification signals (correct or incorrect) can be reliably computed with simple rule-based verifiers \citep{team2025kimi, gandhi2024stream, zhang2024openrft}. Nonetheless, the extension of RLVR to broader, more nuanced domains remains largely unexplored, due primarily to the challenges associated with verifying complex, frequently unstructured reference answers.

In this paper, we aim to extend the applicability of RLVR to domains beyond structured mathematics and coding, by investigating its performance in a diverse set of complex reasoning-intensive areas such as medicine, chemistry, psychology, economics, and education. Central to this exploration is the observation that binary correctness judgments, even on broad-domain tasks, tend to exhibit remarkable agreement across varied large language models (LLMs), including both closed-source models (e.g., GPT-4o) and recently released powerful open-source solutions (e.g., Qwen2.5-72B-Instruct) when provided high-quality objective references authored by domain experts. This finding indicates that reference-based evaluation of diverse domain answers is typically easier than reference-free verification, which is inherently as difficult as identifying the first mistake in a response~\citep{lightman2023let}. Consequently, this insight undermines the presumed necessity for extensive domain-specific annotation and motivates rethinking traditional practices in reward-model training for multi-domain scenarios.

While binary rewards have been the prevalent standard across RLVR applications \citep{gandhi2024stream, lambert2024t, guo2025deepseek, ma2025s}, they pose clear limitations—especially for unstructured tasks. Notably, our data analysis on real-world exam questions reveals that only 60.3\% of mathematical problems possess single-term numerical answers verifiable by rule-based methods, with the ratio dropping further to 45.4\% for complex multi-domain queries. This presents inherent challenges for binary reward schemes and demonstrates the need for richer and more granular verification mechanisms.
To address these limitations, we propose incorporating soft scores obtained from generative, model-based verifiers directly into RLVR. Specifically, we compute a soft reward from the probability of a single indicative token produced by a generative verifier summarizing its assessment.   Crucially, we demonstrate that it is feasible to distill effective multi-domain generative verifier models based on relatively compact models (sizes as small as 7B) without conducting extensive domain-specific annotation. Instead, we employ data composed of response samples and their corresponding judgments collected during RL exploration under the supervision of a larger cross-domain generative teacher model. These noisy yet more realistic datasets promote robustness of the subsequently distilled model-based rewards.

Our empirical results strongly validate the effectiveness of our extended RLVR framework across various domains. By fine-tuning modest-sized (7B) base models using various RL algorithms and our soft reward verifier, we obtain improved reasoning policies superior to state-of-the-art open-source alignment models such as Qwen2.5-72B-Instruct and DeepSeek-R1-Distill-Qwen-32B, achieving performance boosts of up to 8.0\%  accuracy in diverse, free-form reasoning tasks. We particularly observe that our model-based soft rewards consistently scale better and produce more robust policies compared to conventional rule-based binary rewards, especially on unstructured answer scenarios and larger training data regimes.

\textbf{Contributions.} Our key contributions can be summarized as follows:
\begin{itemize}[leftmargin=12pt]
  \item We extend reinforcement learning with verifiable rewards (RLVR) to diverse domains, establishing its effectiveness beyond traditional structured answer scenarios.
  \item We introduce and validate a novel framework incorporating generative model-based soft rewards within RLVR, demonstrating substantial improvements in generalization, robustness, and scalability relative to traditional binary rule-based rewards.
  \item We empirically demonstrate the feasibility and efficacy of training compact (7B-scale) cross-domain generative reward verifiers without extensive domain-specific annotation, challenging traditional assumptions about annotation scale.
  \item We release a dataset containing 570k examples of multi-domain free-form data and the corresponding trained reward model, available at \url{https://huggingface.co/collections/virtuoussy/rlvr-67ea349b086e3511f86d1c1f}, to facilitate future research in this promising direction.
\end{itemize}

\section{Related Work}

\subsection{Reward Estimation in Reinforcement Learning with Verifiable Rewards}

For reasoning tasks such as mathematical reasoning, whether in constructing training data or at test time, a solution is typically considered correct if it arrives at correct final answer~\citep{cobbe2021training}. This is because reliably assessing the correctness of individual steps remains an open challenge, particularly when these steps may lack ground-truth labels in real-world scenarios. Similarly, the correctness of solutions to coding problems is typically accessed based on whether all test cases pass~\citep{austin2021program,hendrycks2021measuring,gehring2024rlef}. Consequently, previous reference-based RL studies have primarily focused on mathematical reasoning and coding tasks.

In most previous studies~\citep{zelikman2022star,gandhi2024stream,zhang2024openrft,lambert2024t,guo2025deepseek,ma2025s,yu2025dapo}, given access to the reference answer $a$, the correctness label $z$ for a response $y$ to a prompt $x$ is typically a binary value. $z$ can also take on a value in the range \([0, 1]\) to reflect varying degrees of correctness~\citep{luong2024reft,li2024humans,ma2025sorft,xie2025logic}. Labels are assigned by a deterministic function $z=f(x,y,a)$, which operates based on predefined rules (e.g., exact match). These rules can also be combined with tools, such as a Python library, for verification~\citep{xiong2025selfrewarding,deepscaler2025}. This method is particularly effective when the answer type is fixed and easily matchable, such as a numerical value or a multiple-choice option. Each response is rated individually, without considering any preference information.

Besides using closed-source LLMs such as GPT-4o as verifiers~\citep{chen2024huatuogpt}, recent studies have also explored training reference-based reward models for mathematical reasoning~\citep{team2025kimi}. However, these models are confined to a single domain and still require large-scale training data (e.g., $800$k instances for math) even within that domain.

\subsection{Generative Reward Modeling}

Using next-token prediction for reward modeling has attracted great interest in recent years~\citep{lightman2023let,zheng2023judging,NEURIPS2024_5e5853f3,zhang2024generative}, as it enables LLMs to fully leverage their generative capabilities, not only to produce accurate rewards but also to provide rationales that justify their judgments. In this work, we explore applying generative, reference-based verifiers to reinforcement learning and investigate their effectiveness across a variety of domains, an area that remains largely underexplored.

Furthermore, we explore training generative reward models without the need for annotated or synthetic step-by-step rationales~\citep{team2025kimi,zhang2024generative} to justify the final assessment. Specifically, we leverage the confidence of generative verifiers to provide stable and informative reward signals, enhancing the robustness of RL training in the presence of noise and ambiguity.

\subsection{Verifiable Reasoning Data}

Previous and on-going RLVR studies primarily focus on narrow tasks~\citep{code-r1,xie2025logic} such as math word problem solving, code generation, and logic puzzles, where well-structured reference answers allow for straightforward rule-based verification. For example, SimpleRL~\citep{zeng2025simplerlzooinvestigatingtamingzero} and Tulu~\citep{lambert2024t} use math datasets GSM8K~\citep{gsm8k} and MATH~\citep{math}, in which each reference answer typically consists of fewer than two words. However, this reliance on well-structured data constrains the scale and diversity of resources that can be used for RLVR across broader domains. 

In this work, we explore RLVR using reasoning data spanning diverse domains, where reference answers are free-form, either written by domain experts for unbiased evaluation~\citep{yu-2021-self-teaching}, extracted from pre-training corpora~\citep{yue2024mammoth2}, or generated by LLMs~\citep{yuan2025naturalreasoning}.

\section{Method}

We focus on a setting where each prompt $x$ is accompanied by an expert-written reference answer $a$. Reference answers have been shown to play a crucial role in providing accurate rewards for reinforcement learning in reasoning-intensive tasks such as coding and mathematics~\citep{shao2024deepseekmath}. Ideally, in these domains, a response $y$ can be objectively verified against the given reference answer $a$. However, in practice, this verification process may be influenced by factors such as imperfect answer extraction and matching when pattern-based verifiers are used, as well as noise introduced by automated evaluation systems, such as a reward model $r_{\phi}(x, a, y)$.

Nevertheless, we can still use this verifiable reward in a policy gradient algorithm, with REINFORCE~\citep{williams1992simple} as an example, as follows:

\begin{equation}
J(\theta) = \E_{(x,a) \sim D} \E_{y_i \sim \pi_{\theta}(\cdot \mid x)} \big[ r_{\phi}(x, a, y_i) \Big].
\end{equation}

When the generation of an entire response is modeled as a single action~\citep{ahmadian2024back}, the gradient becomes (see Section~\ref{appendix:reinforce} for details):

\begin{equation}
\nabla_{\theta} J(\theta) = \E_{(x,a) \sim D} \E_{y_i \sim \pi_{\theta}(\cdot \mid x)} \big[ r_{\phi}(x, a, y_i) \nabla_{\theta} \log \pi_{\theta}(y_i \mid x)  \big].
\end{equation}

\subsection{Reward Estimation}
\label{sec:method:estimation}

To ensure a binary reward signal, we instruct a generative LLM $\pi_{\phi}$ to output only $0$ or $1$ (see system prompt in Table~\ref{tab:appendix:grade_template}). For notational simplicity, we assume that each response consists of exactly $T$ steps, where each step corresponds to a non-empty line. Let $y_i^T$ denote the final step of response $y_i$. The binary model-based reward function is then defined as:

\begin{equation}
r_{\phi}(x, a, y_i) = \mathbbm{1} \big( c_i  = 1 \big),
\label{eq:binary}
\end{equation}

\indent where $c_i$ is sampled from $\pi_{\phi}(\cdot \mid  x, a, y_i^T)$, representing $\pi_{\phi}$'s judgment on the correctness of $y_i$.

Using $\pi_{\phi}$ as a verifier, we can also define a soft reward function using the probability of the judgment tokens (i.e., $0$ or $1$):

\begin{equation}
r_{\phi}(x, a, y_i) =
\begin{cases} 
    \pi_{\phi}(1 \mid x, a, y_i^T) & \text{if } c_i = 1, \\
    1 - \pi_{\phi}(0 \mid x, a, y_i^T) & \text{if } c_i = 0, \\
    0 & \text{otherwise}.
\end{cases}
\label{eq:prob}
\end{equation}

As shown in Equations~\ref{eq:binary} and~\ref{eq:prob}, $r_{\phi}(x, a, y_i)$ is bounded within $[0, 1]$, ensuring consistency with the widely adopted binary reward scale.

\subsection{Reward Normalization}
\label{sec:method:normalization}

To ensure stable gradients and encourage improvement across all samples in a batch that perform above average, we apply z-score normalization to rewards, inspired by prior studies such as GRPO~\citep{shao2024deepseekmath} and REINFORCE++~\citep{hu2025reinforce++}. 

\begin{equation}
\tilde{r}(x, a, y_i) = \frac{r(x, a, y_i)-\mu_r}{\sigma_r},
\end{equation}

where $\mu_r$ and $\sigma_r$ denote the mean and standard deviation of the rewards within the batch containing $y_i$, respectively. In the special case where $\sigma_r=0$, we set all normalized rewards to zero, as these samples are either too difficult or too easy for the current policy.

\subsection{Reward Model Training}
\label{sec:method:training}

When considering generative verifiers, a natural choice is to use an off-the-shelf aligned LLM as the reward model $\pi_{\phi}$, inspired by prior work that employs LLMs as judges~\citep{zheng2023judging}. However, we observe a noticeable performance gap on downstream tasks when using LLMs of different sizes. For example, the 72B reward model achieves 62.7\% while the 7B model gets 58.8\% on math data (see training details in Section \ref{sec:experiments}). To address this, we explore training a moderately sized reward model (e.g., 7B) for general domains, aiming to balance performance and efficiency.

Since there are no ground-truth reward labels, for each $(x, a, y)$ triple, we prompt a fixed LLM to obtain the binary judgments $c \in \{0, 1\}$, indicating whether $y$ matches the reference answer $a$.
During the RL phase, we collect the data $\{(x, a, y, c)\}$ from the exploration stages and use it to fine-tune our reward models with supervised learning on $c$. Unlike relying on a fixed LLM to generate $y$, the improving actor policy produces responses with varying performance and potential formatting noise, which may enhance the robustness of the trained reward models.

\section{Experiments}
\label{sec:experiments}
\subsection{Data}
\label{sec:experiments:data}
\paragraph{Mathematics Data}
To ensure high-quality reference answers, we use a large-scale dataset of 773k Chinese Question Answering (QA) pairs, collected under authorized licenses from educational websites. This dataset covers three educational levels: elementary, middle, and high school. Unlike well-structured yet small-scale benchmarks such as MATH~\citep{math} and GSM8K~\citep{gsm8k}, our reference answers are inherently free-form, often interwoven with rationales or involving several sub-questions yet lacking clear structural patterns. As a result, rule-based reward functions that rely on clean, well-structured answers for verification struggle to process these unstructured reference answers effectively. 

We use GPT-4o-mini to translate questions and their corresponding responses into English. We randomly sample 3,000 QA pairs from each level and reserve them for testing. The average length of reference answers in the test set is 33.7, 36.3, and 53.9 words for elementary, middle, and high school levels, respectively. These are much longer than those in the GSM8K (1 word) and MATH (1.3 words) test sets. 

\paragraph{Multi-Subject Data}
Since no large-scale, free-form dataset with objective reference answers exists for general domains, we use a multi-subject multiple-choice QA dataset ExamQA~\citep{yu-2021-self-teaching}. Originally written in Chinese, ExamQA covers at least 48 first-level subjects. We remove the distractors and convert each instance into a free-form QA pair. This dataset consists of 638k \textbf{college-level} instances, with both questions and objective answers written by domain experts for examination purposes. We also use GPT-4o-mini to translate questions and options into English.

\begin{figure*}[h!]
   \begin{center}
   \includegraphics[width=0.9\textwidth]{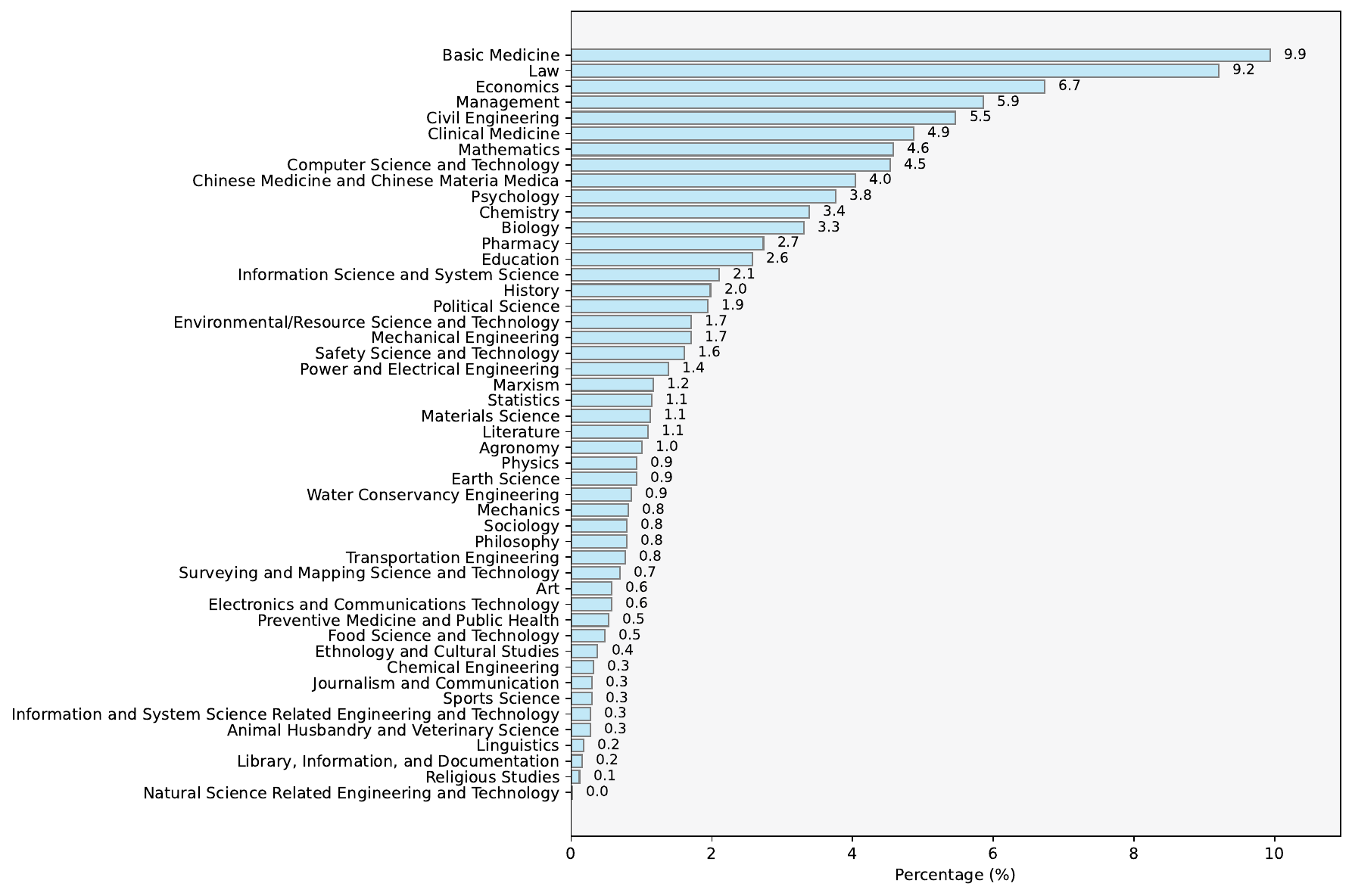}
   \end{center}
 \caption{Distribution of subject occurrences in the test set of ExamQA (excluding unclassified).}
 \label{fig:examqa}
\end{figure*}

For evaluation, we randomly sample 6,000 questions from ExamQA as the test set, while the remaining questions are used as the training pool. Since subject labels are not provided for each QA pair, we use GPT-4o-mini to classify them into one of 48 subjects or mark them as unclassified if uncertain. The detailed classification prompt is provided in Table~\ref{tab:appendix:classification_template}. Excluding unclassified instances (15.8\% of the test data), the most frequent subjects include basic medicine, law, economics, management, civil engineering, mathematics, computer science and technology, psychology, and chemistry, as shown in Figure~\ref{fig:examqa}. For ease of analysis, we further categorize these subjects into four broad fields (STEM, social sciences, humanities, and applied sciences) as detailed in Table~\ref{tab:appendix:four_types}. See examples in Table~\ref{tab:appendix:examqa_examples}.

\paragraph{Data for Training the Reward Model}
We construct the data for training the reward model by extracting 20k samples from each training set of the two datasets, totaling 40k samples. Using the methodology in Section \ref{sec:method:training}, we employ Qwen2.5-7B~\citep{qwen2.5} to conduct RL training. We use the RLOO~\citep{kool2019buy,ahmadian2024back} algorithm and generate four online samples for each prompt. We use Qwen2.5-72B-Instruct as the reward model for hard label determination. By preserving all input-output pairs, this process yields 160k distilled training samples from Qwen2.5-72B-Instruct for reward model training.

To verify the training approach's validity, we exclude these 40k original samples from the final training dataset. This strict separation ensures that the reward model never encounters any data used in previous training stages, thereby guaranteeing evaluation objectivity.

\subsection{Baselines and Notations}
\noindent \textbf{Base} \quad Directly use the base model to generate the response of the question.

\noindent \textbf{SFT} \quad Directly use the label (without CoT) to fine-tune the base model.

\noindent \textbf{Rule-based reward} \quad RL with the reward determined by predefined rules.

\noindent \textbf{Qwen2.5-72B-Instruct} \quad RL with the reward determined by the judgment of Qwen2.5-72B-Instruct~\citep{qwen2.5}.

\noindent \textbf{RM-7B (ours)} \quad RL with the reward determined by the judgment of the reward model trained on our 160k distilled data based on Qwen2.5-7B-Instruct~\citep{qwen2.5}.

\noindent \textbf{Binary} \quad When using rule-based rewards, we directly judge if the label is in the answer. When using model-based rewards, we use the output of the model. The value of binary reward should be in $\{0,1\}$.

\noindent \textbf{Soft} \quad When using rule-based rewards, we use Jaccard similarity ~\citep{jaccard} as the reward.  When using model-based rewards, we use the probability of the first output token. The value of soft reward should be in $[0,1]$.
\subsection{Evaluation}

We begin by investigating majority voting using a strong open-source LLM, Qwen2.5-72B-Instruct~\citep{qwen2.5}, as the reward model $\pi_{\phi}$. The evaluation process follows the prompting template provided in Table~\ref{tab:appendix:grade_template}. Given a prompt $x$ and a reference answer $a$, we generate $m$ evaluation samples and determine the correctness of a response $y$ via majority voting. A response is considered correct if at least half of the evaluations classify it as such, i.e., $\sum_{j=1}^{m} \mathbbm{1} \big[  \pi_{\phi}^{(j)} (x, y^T, a) = 1  \big] \geq \frac{m}{2}$. 

We measure the agreement between the Qwen-based evaluation method (majority voting over $m$ samples) and GPT-4o (a single evaluation per response) using Cohen’s Kappa ($\kappa$). As shown in Figure~\ref{fig:agreement}, the two evaluation methods demonstrate almost perfect agreement (\(0.81 \leq \kappa \leq 1.00\)), with $\kappa$ exceeding $0.86$ for mathematics and $0.88$ for multi-subject college-level problems. This high level of agreement remains consistent across varying values of $m$, indicating that the results are not highly sensitive to the number of evaluation samples. Based on this observation, we adopt $m=1$ in all subsequent evaluations to improve efficiency without compromising evaluation quality.

\subsection{Implementation Details}
After obtaining the 160k distilled data from Qwen2.5-72B-Instruct, we perform supervised fine-tuning on Qwen2.5-7B-Instruct using this data, resulting in our reward model.
We use different RL algorithms to validate the effectiveness of our method, including REINFORCE~\citep{williams1992simple,ahmadian2024back}, RLOO~\citep{kool2019buy,ahmadian2024back}, and REINFORCE++~\citep{hu2025reinforce++}.
Following~\citet{stiennon2020learning, ouyang2022training, hu2025reinforce++}, we introduce a Kullback-Leibler (KL) divergence penalty between the RL model and the reference policy (i.e., base model) distributions to mitigate bias in the reward model. We update $\tilde{r}(x, a, y_i)$ as follows:

\begin{equation}
\tilde{r}(x, a, y_i) \gets \tilde{r}(x, a, y_i) - \beta \log \Bigg( \frac{\pi_{\theta}(y_i \mid x)}{\pi_{\text{ref}}(y_i \mid x)}\Bigg),
\end{equation}

\noindent where $\beta \geq 0$ controls the effect of the KL penalty, and $\pi_{\text{ref}}$ represents the reference policy distribution. We set $\beta=0.01$ for all experiments.

For all algorithms, we apply reward normalization as introduced in Section~\ref{sec:method:normalization}.
We use Qwen2.5-7B~\citep{qwen2.5} as the base model for our experiments. Despite not undergoing post-training, it demonstrates reasonable instruction-following capabilities, as shown by its zero-shot performance in Table~\ref{tab:main_results}. 
We also include the results of Qwen2.5-72B-Instruct and DeepSeek-R1-Distill-Qwen-32B to illustrate the difficulty level of our datasets.
For both datasets, we select 30k samples as the training data.
The training hyper-parameters of RL distilled data collection, reward model training, and the main experiments can be found in Table \ref{tab:appendix:hyper parameters} in the Appendix.

\begin{table}[ht!]
\centering
\small
\resizebox{\textwidth}{!}{
\begin{tabular}{lllcccccccccc}
\toprule
\multirow{2}{*}{\textbf{Method}} & \multirow{2}{*}{\textbf{Reward}} & \multirow{2}{*}{\textbf{Score Type}} & \multicolumn{4}{c}{\textbf{Math}} & \multicolumn{6}{c}{\textbf{Multi-Subject}} \\
\cmidrule(lr){4-7} \cmidrule(lr){8-13} 
 & & & \textbf{E} & \textbf{M} & \textbf{H} &\textbf{Avg}
 &\textbf{STEM}&\textbf{Social}&\textbf{Humanities}&\textbf{Applied}&\textbf{Others}&\textbf{Avg}
 \\
\midrule

Qwen2.5-72B-Instruct & -- &-- & 44.2 & 57.7 & 40.3 & 47.4 & 25.2 &20.1 & 28.7 & 20.5 &21.0 & 22.6 \\ 

DeepSeek-R1-Distill-Qwen-32B & -- & -- & 27.6 & 34.8 & 17.4 & 26.6 & 23.2 & 21.8 & 26.7 & 20.5 & 18.5 & 21.7   \\ 

\midrule
Base & -- &-- & 43.1 & 53.9 & 33.2 & 43.4 & 16.3 & 14.9 & 15.2 & 13.3 & 14.8 & 15.0 \\ 
\midrule

SFT & -- &-- & 53.6 & 50.5 & 32.9 & 45.7 & 24.6 & 22.8 & 25.7 & 20.9 & 22.6 & 23.1 \\ 
\midrule
\multirow{6}{*}{REINFORCE}
&\multirow{2}{*}{rule based}& binary & 58.5 & 66.5 & 46.7 & 57.2 & 25.3 & 26.6 & 27.7 & 21.1 & 20.7 & 24.2 \\ && soft & 46.0 & 47.7 & 31.5 & 41.7 & 22.0 & 20.3 & 23.1 & 16.9 & 20.5 & 20.3 \\
&\multirow{2}{*}{Qwen2.5-72B\textsubscript{Instruct}}& binary & \textbf{64.4} & \textbf{72.1} & 51.6 & \textbf{62.7} & 27.9 & 27.9 & 30.7 & 24.4 & 23.2 & 26.6 \\ && soft & 62.5 & 71.2 & \textbf{53.1} & 62.3 & 32.2 & 32.8 & \textbf{36.0} & 24.9 & \textbf{27.9} & 30.3 \\
&\multirow{2}{*}{RM-7B (ours)}& binary & 63.8 & 71.7 & 51.9 & 62.5 & 29.0 & 29.1 & 28.4 & 23.8 & 24.8 & 27.3 \\ && soft & 62.9 & 70.7 & 53.0 & 62.2 & \textbf{32.7} & \textbf{32.8} & 35.6 & \textbf{28.6} & 27.4 & \textbf{31.2} \\

\midrule

\multirow{6}{*}{REINFORCE++}
&\multirow{2}{*}{rule based}& binary & 56.4 & 65.5 & 47.6 & 56.5 & 26.1 & 26.1 & 26.4 & 21.8 & 24.7 & 25.0 \\ 
&& soft & 49.4 & 52.9 & 36.2 & 46.2 & 22.5 & 22.0 & 25.7 & 18.6 & 20.2 & 21.4 \\
&\multirow{2}{*}{Qwen2.5-72B\textsubscript{Instruct}}& binary & 63.0 & 71.3 & 50.4 & 61.6 & 30.7 & \textbf{32.8} & \textbf{34.3} & \textbf{27.5} & \textbf{27.8} & \textbf{30.3} \\ && soft & 62.7 & 70.4 & 50.5 & 61.2 & \textbf{30.8} & 30.1 & 33.7 & 25.6 & 25.4 & 28.8 \\
&\multirow{2}{*}{RM-7B (ours)}& binary & \textbf{63.1} & \textbf{71.3} & \textbf{51.5} & \textbf{62.0} & 30.2 & 30.8 & 31.0 & 26.6 & 26.3 & 29.1 \\ && soft & 62.7 & 70.3 & 50.8 & 61.3 & 29.5 & 31.7 & 33.7 & 25.8 & 26.2 & 29.0 \\
\midrule
\multirow{6}{*}{RLOO}
&\multirow{2}{*}{rule based}& binary & 58.2 & 67.0 & 50.2 & 58.5 & 28.2 & 27.9 & 27.4 & 22.4 & 24.5 & 26.3 \\ && soft & 49.6 & 50.3 & 33.9 & 44.6 & 16.7 & 17.3 & 18.8 & 14.5 & 16.9 & 16.6 \\
&\multirow{2}{*}{Qwen2.5-72B\textsubscript{Instruct}}& binary & 63.0 & 70.8 & 51.1 & 61.6 & 29.4 & 30.5 & 33.7 & 24.6 & 26.1 & 28.4 \\ 
&& soft & \textbf{63.8} & 71.0 & 52.4 & 62.4 & \textbf{32.9} & 31.4 & 34.7 & \textbf{27.7} & 26.8 & \textbf{30.6} \\
&\multirow{2}{*}{RM-7B (ours)}& binary & 63.4 & \textbf{71.8} & \textbf{53.8} & \textbf{63.0} & 29.3 & 29.0 & 33.3 & 25.8 & 25.6 & 28.1 \\ 
&& soft & 63.3 & 71.7 & 53.6 & 62.9 & 31.0 & \textbf{32.0} & \textbf{35.6} & 27.0 & \textbf{27.0} & 30.0 \\
\midrule
\bottomrule
\end{tabular}
}
\hspace{0.1pt}
\caption{Performance Comparison of Different Methods. Base model: Qwen2.5-7B. E: elementary. M: middle. H: high.}
\label{tab:main_results}
\end{table}

\subsection{Main Results}
Table \ref{tab:main_results} shows the results on mathematics and multi-subject tasks.
We have the following observations:

\noindent\textbf{Evaluation on Base Models} \quad
Both our math and multi-subject data have demonstrated notable difficulty, with even strong open-source models like Qwen2.5-72B-Instruct~\citep{qwen2.5} and DeepSeek-R1-Distill-Qwen-32B~\citep{guo2025deepseek} performing unsatisfactorily, particularly on multi-subject tasks (21.7\% for DeepSeek-R1-Distill-Qwen-32B and 22.6\% for Qwen2.5-72B-Instruct). We believe that more challenging datasets will better facilitate exploration across the industry.

\noindent\textbf{SFT vs. RL} \quad
SFT significantly underperforms RL on both math and multi-subject tasks. Notably on math, SFT merely improves the model performance from 43.4\% to 45.7\%, falling far short of rule-based reward RL (RLOO, 58.8\%) and lagging even further behind model-based reward RL (RM-7B, 63.0\%). These findings demonstrate RL's distinct advantages and potential in reasoning tasks when there is no high-quality Chain-of-Thoughts for training.

\noindent\textbf{Model-based Reward vs. Rule-based Reward} \quad
From the table, we can conclude that model-based reward consistently outperforms rule-based reward in free-form reference-based scenarios. For instance, RM-7B (ours) and Qwen-2.5-72b-Instruct with binary reward achieves 63.0\% and 61.6\% respectively on average with RLOO, while rule-based reward only gets 58.5\%.
Notably, our distilled 7B reward model exhibits competitive performance against its much larger predecessor, Qwen2.5-72B-Instruct. In multi-subject evaluations using REINFORCE, the model trained from RM-7B achieves 31.2\% accuracy compared to the 72B model's 30.3\% – a significant improvement given the substantial parameter disparity. This enhanced capability likely emerges from stabilized response patterns developed during training, which better align with the generative reward model's objectives compared to the base model's more variable outputs.

\noindent\textbf{Binary Reward vs. Soft Reward} \quad
For rule-based reward, soft reward consistently underperforms binary reward. This discrepancy may stem from redundant tokens between the model's generated answers and reference labels, which can lower the reward scores for correct answers. A potential improvement could involve adopting metrics like cosine similarity of sentence embeddings as soft rewards, as these may better capture semantic alignment.
In contrast, for model-based reward, binary and soft rewards yield comparable results on math tasks. This suggests that the model likely produces judgments with extremely high confidence, as determining answer-label matches in mathematical problems is relatively easy. However, in multi-subject tasks, where reference labels exhibit greater diversity and consequently higher judgment complexity, soft rewards demonstrate more conservative scoring behavior. This conservatism in ambiguous cases enables soft rewards to outperform binary rewards in certain scenarios (31.2\% vs. 27.3\%, REINFORCE, RM-7B), as their soft scoring better accommodates the inherent uncertainty of open-domain evaluation.

\noindent\textbf{Summary} \quad
Our method establishes new state-of-the-art performance in RLVR through three key innovations:
(1) Our proposed model-based reward is much stronger than rule-based baseline, allowing various RL methods to obtain very accurate rewards in general domain scenarios.
(2) Building upon the data distilled from Qwen2.5-72B-Instruct, we develop a computationally efficient 7B model that can achieve comparable or even better performance.
(3) We extend binary reward to soft reward, which can get more conservative scores for ambiguous cases, which can help get better performance when the reference answers exhibit greater diversity and consequently higher judgment complexity.

\subsection{Scaling Experiments}
\label{sec:analysis:scaling}
\begin{table}[ht!]
\centering
\small
\resizebox{\textwidth}{!}{
\begin{tabular}{llcccccccccc}
\toprule
\multirow{2}{*}{\textbf{Method}} & \multirow{2}{*}{\textbf{Scale}} & \multicolumn{4}{c}{\textbf{Math}} & \multicolumn{6}{c}{\textbf{Multi-Subject}} \\
\cmidrule(lr){3-6} \cmidrule(lr){7-12} 
 & & \textbf{E} & \textbf{M} & \textbf{H} &\textbf{Avg}
&\textbf{STEM}&\textbf{Social}&\textbf{Humanities}&\textbf{Applied}&\textbf{Others}&\textbf{Avg}
 \\
\midrule

\multirow{7}{*}{Rule based}
&20k &58.9 &68.1 &47.6 &58.2 &27.3 &28.0 &31.4 &23.5 &23.0 &26.2 \\
&40k &61.5 &69.4 &55.4 &62.1 &25.1 &24.8 &27.4 &21.0 &23.0 &24.0 \\
&60k &62.6 &69.8 &56.8 &63.1 &20.0 &21.9 &26.4 &16.6 &19.9 &20.1 \\
&80k &62.4 &68.2 &53.6 &61.4 &19.2 &18.3 &26.7 &15.1 &16.4 &18.0 \\
&100k &52.6 &57.2 &45.2 &51.7 &17.8 &18.2 &20.5 &13.4 &16.4 &16.9 \\
\midrule
\multirow{7}{*}{RM-7B (ours)}
&20k &64.9 &71.8 &53.4 &63.4 &30.8 &34.6 &31.7 &28.0 &27.7 &30.8 \\
&40k &65.6 &72.4 &54.4 &64.1 &34.3 &33.7 &36.3 &29.5 &28.6 &32.4 \\
&60k &66.0 &71.6 &53.2 &63.6 &33.3 &36.6 &37.3 &31.5 &28.9 &33.3 \\
&80k &66.6 &72.3 &55.6 &64.8 &34.5 &38.6 &38.3 &31.6 &31.0 &34.6 \\
&100k &67.1 &72.3 &55.6 &65.0 &35.1 &38.5 &39.3 &32.7 &30.7 &35.0 \\
\midrule
\bottomrule
\end{tabular}
}
\hspace{0.1pt}
\caption{The results of the scaling experiments. We use RLOO as the RL algorithm and binary reward as the score type.}
\label{tab:scaling}
\end{table}
Scalability has emerged as a critical property in the RL-based training era. A key question worthy of investigation is whether model performance can continue to improve as RL training progresses and data volume increases. To examine this, we conduct experiments using our trained reward model against rule-based reward while progressively scaling the dataset. We randomly sampled 100k samples from our training corpus as the baseline set, conducting evaluations on both mathematical reasoning and multi-subject tasks. Table \ref{tab:scaling} shows the experimental results.

The results reveal significant differences in scaling capabilities. The rule-based reward demonstrates unstable scalability across both mathematical and multi-subject tasks, exhibiting substantial performance fluctuations and eventual degradation as RL training continues. In contrast, our learned reward model shows consistent improvement trends throughout the training process. This empirical evidence highlights the inherent scalability advantages of model-based rewards compared to rule-based rewards.

\subsection{Out-of-Distribution Evaluation}
\begin{table}[ht!]
\centering
\small
\begin{tabular}{lcc}
\toprule
Method & Natural Reasoning & WebInstruct \\
\midrule
Rule based &29.4&33.9 \\

\midrule
RM-7B (ours)  &39.8& 44.0 \\

\midrule
\bottomrule
\end{tabular}

\hspace{0.1pt}
\caption{The results of the Out-of-Distribution evaluation}
\label{tab:ood_results}
\end{table}
To further validate the effectiveness of our reward model, we conduct additional evaluations on two benchmarks: NaturalReasoning~\citep{yuan2025naturalreasoning} and WebInstruct~\citep{yue2024mammoth2}. We compare the performance of the rule-based reward with our RM-7B using RLOO with binary reward.
The base model is Qwen2.5-7B.
For both datasets, we randomly select 30K examples for training and 5K sample for evaluation.
The results are shown in Table \ref{tab:ood_results}.
As can be seen from the table, the performance of RM-7B remains significantly superior to the rule-based reward on datasets from other domains. This demonstrates that our general-purpose reward model can extend to other domains while maintaining strong performance.

\section{Discussions and Conclusions}

In this work, we simplify the verification task by instructing a generative reward model to output either $1$ or $0$, without requiring chain-of-thought (CoT) reasoning~\citep{nye2021show,wei2022chain}. While CoT has proven useful in both reference-based~\citep{team2025kimi} and reference-free~\citep{zhang2024generative} settings, it remains an open question how necessary in-depth rationales are for assessing semantic equivalence between reference answers and model responses in the same language, particularly when focusing on the conclusive part of each response. This also raises a related question for process reward modeling~\citep{lightman2023let} in RLVR: how should rewards be assigned when there is no direct supervision for intermediate steps, regardless of the step segmentation method?

In addition, we do not consider format-based rewards~\citep{guo2025deepseek,xie2025logic} in this work. We revisit the role of format-related constraints and rewards in this context. In prior work, pattern-based functions are often used for scoring, making it critical to guide LLMs to enclose their final answers in an easily parsed format. These extracted answers are then compared with the reference answers for verification and evaluation. In contrast, by reintroducing a reward model in RLVR without imposing any format constraints on reference answers or model responses, we reduce the need for extensive human effort in data standardization and pattern design.

\bibliography{neurips_2024}
\bibliographystyle{colm2024_conference}

\clearpage
\newpage
\appendix
\section{Appendix}

\subsection{Template}
Table \ref{tab:appendix:grade_template} shows the template for the grading task. Table \ref{tab:appendix:classification_template} shows the template for the classification task. Table \ref{tab:appendix:four_types} shows the classification of subjects into STEM (Science, Technology, Engineering, and Mathematics),
Social Sciences, Humanities, and Applied Sciences.

\begin{table*}[ht!]
\centering
\footnotesize

\begin{tabular}{lp{12cm}}
\toprule

& \begin{lstlisting}[basicstyle=\ttfamily\scriptsize, breaklines=true, aboveskip=0pt, belowskip=0pt]
Given a problem, determine whether the final answer in the provided (incomplete) solution process matches the reference answer.  
The reference answer may be one single option character (e.g., A, B, C, D), a numerical value, an expression, or a list of answers if multiple questions are involved.  
**The reference answer may be in Chinese or another language, but your evaluation should be language-agnostic.**  

Your task:  
- Compare the final output of the solution process with the reference answer.  
- If they **match exactly**, output **YES**.  
- If they **do not match**, output **NO**.  
- If the solution process is unclear, incomplete, or ambiguous, assume it is incorrect and output **NO**.  

Your output must be strictly **'YES'** or **'NO'**, with no additional words, punctuation, or explanation.  

---

**Question:**  
{question}  

**Solution Process (Final Step Only):**  
{response}  

**Reference Answer:**  
{reference}  

**Output:**  
\end{lstlisting} \\
\bottomrule
\end{tabular}
\caption{Template for the grading task.}
\label{tab:appendix:grade_template}
\end{table*}
\begin{table*}[ht!]

\centering
\footnotesize

\begin{tabular}{lp{12cm}}
\toprule

& \begin{lstlisting}[basicstyle=\ttfamily\scriptsize, breaklines=true, aboveskip=0pt, belowskip=0pt]
Based on the content of 'Question' and 'Answer' classify the subject into one of the following categories. 

Return only the corresponding subject ID. If classification is uncertain, return 999.

**Question:**  
{question}  

**Answer:**  
{answer}  

110	Mathematics
120	Information Science and System Science
130	Mechanics
140	Physics
150	Chemistry
170	Earth Science
180	Biology
190	Psychology
210	Agronomy
230	Animal Husbandry and Veterinary Science
310	Basic Medicine
320	Clinical Medicine
330	Preventive Medicine and Public Health
350	Pharmacy
360	Chinese Medicine and Chinese Materia Medica
413	Information and System Science Related Engineering and Technology
416	Natural Science Related Engineering and Technology
420	Surveying and Mapping Science and Technology
430	Materials Science
460	Mechanical Engineering
470	Power and Electrical Engineering
510	Electronics and Communications Technology
520	Computer Science and Technology
530	Chemical Engineering
550	Food Science and Technology
560	Civil Engineering
570	Water Conservancy Engineering
580	Transportation Engineering
610	Environmental/Resource Science and Technology
620	Safety Science and Technology
630	Management
710	Marxism
720	Philosophy
730	Religious Studies
740	Linguistics
750	Literature
760	Art
770	History
790	Economics
810	Political Science
820	Law
840	Sociology
850	Ethnology and Cultural Studies
860	Journalism and Communication
870	Library, Information, and Documentation
880	Education
890	Sports Science
910	Statistics
999	Unclassified
    
\end{lstlisting} \\
\bottomrule
\end{tabular}
\caption{Template for the classification task, with subject names and IDs referenced from~\citep{yu-2021-self-teaching}.}
\label{tab:appendix:classification_template}
\end{table*}
\begin{table}[ht!]
\centering
\footnotesize
\begin{tabular}{lp{8cm}}
\toprule
\textbf{Category} & \textbf{Subject IDs} \\
\midrule

\textbf{STEM} & 
\begin{lstlisting}[basicstyle=\ttfamily\scriptsize, breaklines=true, aboveskip=0pt, belowskip=0pt]
110 (Mathematics), 120 (Information Science and System Science), 
130 (Mechanics), 140 (Physics), 150 (Chemistry), 170 (Earth Science), 
180 (Biology), 430 (Materials Science), 460 (Mechanical Engineering), 
470 (Power and Electrical Engineering), 510 (Electronics and Communications Technology), 
520 (Computer Science and Technology), 530 (Chemical Engineering), 
560 (Civil Engineering), 570 (Water Conservancy Engineering), 
580 (Transportation Engineering), 610 (Environmental/Resource Science and Technology), 
620 (Safety Science and Technology), 910 (Statistics) 
\end{lstlisting} \\

\midrule
\textbf{Social Sciences} & 
\begin{lstlisting}[basicstyle=\ttfamily\scriptsize, breaklines=true, aboveskip=0pt, belowskip=0pt]
190 (Psychology), 790 (Economics), 810 (Political Science), 
820 (Law), 840 (Sociology), 850 (Ethnology and Cultural Studies), 
860 (Journalism and Communication), 870 (Library, Information, and Documentation), 
880 (Education), 890 (Sports Science), 630 (Management) 
\end{lstlisting} \\

\midrule
\textbf{Humanities} & 
\begin{lstlisting}[basicstyle=\ttfamily\scriptsize, breaklines=true, aboveskip=0pt, belowskip=0pt]
710 (Marxism), 720 (Philosophy), 730 (Religious Studies), 
740 (Linguistics), 750 (Literature), 760 (Art), 770 (History) 
\end{lstlisting} \\

\midrule
\textbf{Applied Sciences} & 
\begin{lstlisting}[basicstyle=\ttfamily\scriptsize, breaklines=true, aboveskip=0pt, belowskip=0pt]
210 (Agronomy), 230 (Animal Husbandry and Veterinary Science), 
310 (Basic Medicine), 320 (Clinical Medicine), 
330 (Preventive Medicine and Public Health), 350 (Pharmacy), 
360 (Chinese Medicine and Chinese Materia Medica), 
413 (Information and System Science Related Engineering and Technology), 
416 (Natural Science Related Engineering and Technology), 
420 (Surveying and Mapping Science and Technology), 550 (Food Science and Technology) 
\end{lstlisting} \\

\bottomrule
\end{tabular}
\caption{Classification of subjects into STEM (Science, Technology, Engineering, and Mathematics), Social Sciences, Humanities, and Applied Sciences.}
\label{tab:appendix:four_types}
\end{table}

\subsection{Agreement}

\begin{table}[h!]
\centering

\begin{tabular}{lcc}
\toprule

Level & \multicolumn{2}{c}{Agreement (\(\kappa \uparrow\))} \\
\cmidrule(lr){2-3}
 & \(m = 1\) & \(m = 10\) \\
\midrule
elementary & 0.844  &  0.838\\
middle & 0.885 &  0.883\\
high &  0.849 &  0.846\\
average & 0.864 &  0.861 \\
\midrule
\bottomrule
\end{tabular}
\caption{Cohen’s Kappa agreement ($\kappa$) between GPT-4o and majority voting ($m$: the number of votes) using Qwen2.5-72B-Instruct as evaluator across different education levels of math problems.}
\label{tab:experiment:agreement}
\end{table}

Note that for each instance, we have only a single decision from GPT-4o. While it may align more closely with an individual sampled decision from the reward model than with the majority vote (when $m > 1$), the latter provides a more stable and deterministic outcome by reducing randomness during grading. 

\begin{figure*}[h!]
   \begin{center}
   \includegraphics[width=0.8\textwidth]{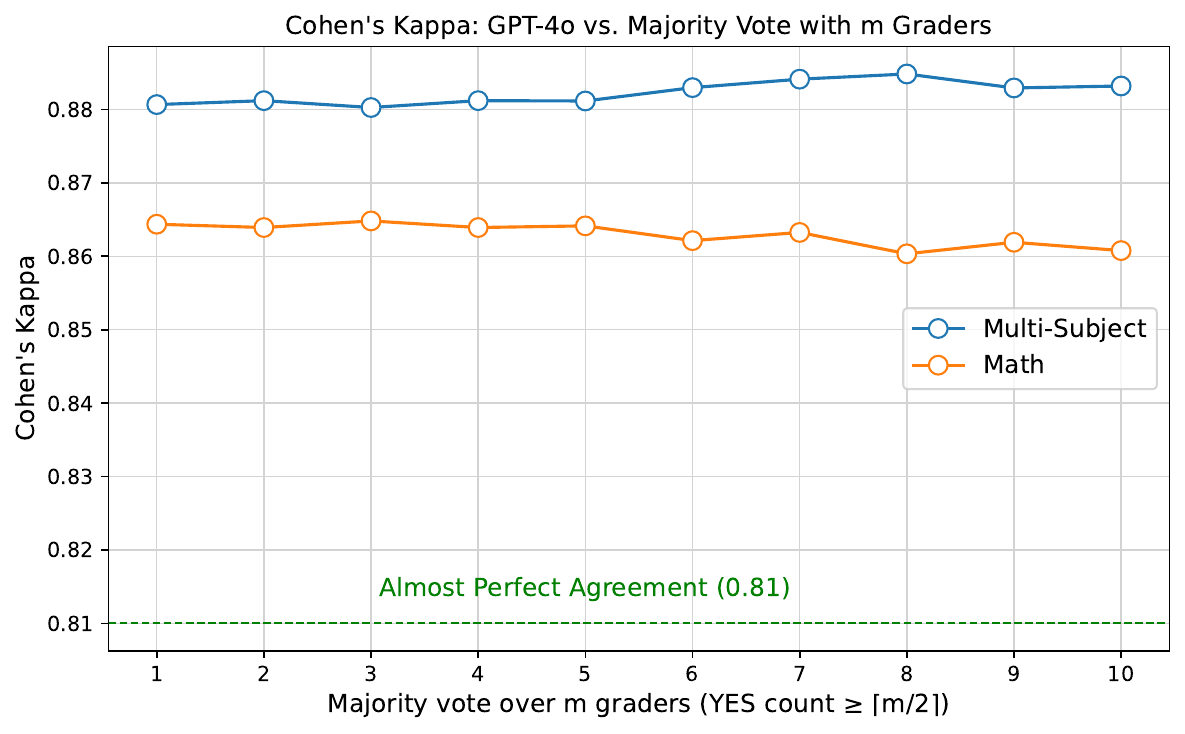}
   \end{center}
 \caption{Agreement between GPT-4o and Majority Vote with m Graders, measured by Cohen's Kappa.}
 \label{fig:agreement}
\end{figure*}

\begin{table}[h!]
\centering

\begin{tabular}{lcc}
\toprule

Level & \multicolumn{2}{c}{Agreement (\(\kappa \uparrow\))} \\
\cmidrule(lr){2-3}
 & \(m = 1\) & \(m = 10\) \\
\midrule
college-level & 0.881  & 0.883 \\
\midrule
\bottomrule
\end{tabular}
\caption{Cohen’s Kappa agreement ($\kappa$) between GPT-4o and majority voting ($m$: the number of votes) using Qwen2.5-72B-Instruct as evaluator across college-level multi-subject problems.}
\label{tab:experiment:agreement2}
\end{table}

\subsection{REINFORCE}
\label{appendix:reinforce}

\begin{equation}
    \begin{aligned}
        \nabla_{\theta} \mathbb{E}_{y_i \sim \pi_{\theta} (\cdot \mid x)} \Big[ r(x, a, y_i) \Big]
        &= \sum_{y_i} \nabla_{\theta} \Big[ \pi_{\theta} (y | x) \Big] r(x, a, y_i) \\
        & = \sum_{y_i} \Big[\pi_{\theta} (y | x) \nabla_{\theta} \log \pi_{\theta} (y | x) \Big] r(x, a, y_i) \\
        &= \mathbb{E}_{y_i \sim \pi_{\theta} (\cdot \mid x)} \Big[ \nabla_{\theta} \log \pi_{\theta} (y_i| x) r(x, a, y_i) \Big].
    \end{aligned}
\end{equation}

\subsection{Hyper parameters}
\begin{table*}[ht!]
\centering
\begin{tabular}{lcccc}
\toprule
\multirow{2}{*}{\textbf{Hyperparameter}} & \multicolumn{2}{c}{\textbf{Reward Training}} & \multicolumn{2}{c}{\textbf{Main Experiments}} \\
\cmidrule(lr){2-3} \cmidrule(lr){4-5}
 & \textbf{RL} & \textbf{SFT} & \textbf{RL} & \textbf{SFT} \\
\midrule
micro\_train\_batch\_size & 8 & 4 & 8 & 4 \\
train\_batch\_size & 128 & 128 & 128 & 128 \\
micro\_rollout\_batch\_size & 16 & -- & 16 & -- \\
rollout\_batch\_size & 128 & -- & 128 & -- \\
n\_samples\_per\_prompt & 4 & -- & 4 & -- \\
max\_samples & 40000 & 1600000 & 30000 & 30000 \\
max\_epochs & 1 & 1 & 1 & 1 \\
prompt\_max\_len & 1024 & -- & 1024 & -- \\
generate\_max\_len & 1024 & -- & 1024 & -- \\
max\_len & -- & 4096 & -- & 4096 \\
actor\_learning\_rate & 5e-7 & -- & 5e-7 & -- \\
init\_kl\_coef & 0.01 & -- & 0.01 & -- \\
\bottomrule
\end{tabular}
\caption{Training hyper parameters. Other hyper parameters are the default configuration in OpenRLHF.}
\label{tab:appendix:hyper parameters}
\end{table*}
Table \ref{tab:appendix:hyper parameters} shows the hyper parameters of our experiments.

\begin{table}[h!]
\centering
\small
\begin{tabular}{lp{2cm}p{5cm}p{4cm}}
\toprule
coarse & fine & question & answer \\
\midrule
Social Sciences             &    Psychology         &    Setting up an activity for students to 'bomb' each other with compliments belongs to ( ).      & Self-awareness guidance  \\
\midrule
 STEM & Civil Engineering & A gravity retaining wall meets the Rankine earth pressure conditions, $H = 3\,\mathrm{m}$, top width $2\,\mathrm{m}$, bottom width $3\,\mathrm{m}$, fill $c = 0$, $\phi = 30^\circ$, $\gamma = 18.0\,\mathrm{kN/m^3}$, the base friction coefficient is 0.4, the anti-sliding stability safety factor $K_s$ and the anti-tilting stability safety factor $K_t$ are respectively () & 2.67; 1.73 \\
 \midrule
 Humanities & Philosophy & Laozi pointed out in the 'Tao Te Ching', 'Without leaving the door, one knows the world; without peering through the window, one knows the way of heaven. The farther one goes, the less one knows. Therefore, the sage knows without traveling, sees without looking, and achieves without doing.' Laozi's view here  &  denies the decisive role of practice in understanding \\
 \midrule
Applied Sciences & Agronomy & Under light, the physiological processes that can occur in the mesophyll cells and vascular bundle sheath cells of wheat (C3) are & Production of ATP and [H] \\
\bottomrule
\end{tabular}
\caption{Example question and reference answer pairs in ExamQA.}
\label{tab:appendix:examqa_examples}
\end{table}

\end{document}